\documentclass{llncs}
\usepackage{amsmath}
\usepackage{graphicx}
\usepackage[square,numbers]{natbib}
\usepackage{amssymb}
\usepackage{pifont}
\usepackage{multirow}
\usepackage{comment}
\usepackage{longtable,booktabs}
\usepackage{array, makecell}
\makeatletter
\renewcommand\bibsection%
{
  \section*{\refname
    \@mkboth{\MakeUppercase{\refname}}{\MakeUppercase{\refname}}}
}
\makeatother
\begin{document}

\title{On the Benefit of Combining Neural, Statistical and External Features for Fake News Identification}
\author{Gaurav Bhatt\inst{1}, Aman Sharma\inst{1}, Shivam Sharma\inst{1}, Ankush Nagpal\inst{1}, Balasubramanian Raman\inst{1}, and Ankush Mittal\inst{2}}

\institute{Indian Institute of Technology, Roorkee, India,\\
\email{gauravbhatt.cs.iitr@gmail.com}
\and
Graphic Era University, India}

\maketitle              

\begin{abstract}
Identifying the veracity of a news article is an interesting problem while automating this process can be a challenging task. Detection of a news article as fake is still an open question as it is contingent on many factors which the current state-of-the-art models fail to incorporate. In this paper, we explore a subtask to fake news identification, and that is stance detection. Given a news article, the task is to determine the relevance of the body and its claim. We present a novel idea that combines the neural, statistical and external features to provide an efficient solution to this problem. We compute the neural embedding from the deep recurrent model, statistical features from the weighted n-gram bag-of-words model and hand crafted external features with the help of feature engineering heuristics. Finally, using deep neural layer all the features are combined, thereby classifying the headline-body news pair as agree, disagree, discuss, or unrelated. We compare our proposed technique with the current state-of-the-art models on the fake news challenge dataset. Through extensive experiments, we find that the proposed model outperforms all the state-of-the-art techniques including the submissions to the fake news challenge.
\end{abstract}

\section{Introduction}
Fake news being a potential threat towards journalism and public discourse has created a buzz across the internet. With the recent advent of social media platforms such as Facebook and Twitter, it has become easier to propagate any information to the masses within minutes. While the propagation of information is proportional to growth of social media, there has been an aggravation in the authenticity of these news articles. These days it has become a lot easier to mislead the masses using a single Facebook or Twitter fake post. For an instance, in the US presidential election of 2016, the fake news has been cited as the foremost contributing factor that affected the outcome \cite{nyt_president}. 

\begin{table*}
\centering
\resizebox{\columnwidth}{!}{
\begin{tabular}{c|c|c}
    \toprule
    Headline & "Robert Plant Ripped up \$800M Led Zeppelin Reunion Contract" & Stance\\
   \midrule
  Body 1 & Led Zeppelin's Robert Plant turned down 500 MILLION to reform supergroup. & Agree \\
   Body 2 & No, Robert Plant did not rip up an \$800 million deal to get Led Zeppelin back together. & Disagree \\
   Body 3 & Robert Plant reportedly tore up an \$800 million Led Zeppelin reunion   deal. & Discuss\\
   Body 4 & Richard Branson's Virgin Galactic is set to launch SpaceShipTwo today.	& Unrelated\\
    \bottomrule
  \end{tabular}
}
\caption{Headline-body pairs along with their relative stance.}
\label{tab:sick}
\end{table*}

The root cause of this problem lies in the fact that none of the social networking sites use any automatic system that can identify the veracity of news flowing across these platforms. A possible reason for this failure is the open domain nature of the problem that adds to the intricacies. The recently organized Fake News Challenge (FNC-1) \cite{fnc} is an initiative in this direction. The aim of this challenge is to build an automatic system that has the capability to identify whether a news article is fake or not. More specifically, given a news article the task is to evaluate the relatedness of the news body towards its headline. The relatedness or stance is the relative perspective of a news article towards a relative claim (shown in Table 1).

The idea behind building a countermeasure for fake news is to use machine learning and natural language processing (NLP) tools that can compute semantic and contextual similarity between the headline and the body, and classify the pairs into one of four categories. Deep learning models have been efficacious in solving many NLP problems that share similarities to fake news which includes but not limited to - computing semantic similarity between sentences \cite{sick,kiros2015skip}, community based question answering \cite{QA1,yang2015wikiqa}, etc. The basic building blocks of all deep models are recurrent networks such as recurrent neural networks (RNN) \cite{rnn_sim}, long short-term memory networks (LSTM) \cite{lstm} and gated recurrent units (GRU) \cite{gru}, and convolution networks such as convolution neural networks (CNN) \cite{cnn}. A deep architecture encodes the given sequence of words into fixed length vector representation which can be used to score the relevance of two textual entities, in our case, relevance of each headline-body pair.

Statistical information related to text can be encoded to vectors using the traditional bag-of-words (BOW) approach. The BOW approaches are often combined with term frequency (TF) and inverse document frequency (IDF), and n-grams that helps to encode more information related to the text \cite{riedel2017simple,davisfake_stanford}. These approaches, however simple, have been used to ameliorate the performance of deep models in complex NLP problems such as community question answering \cite{yang2015wikiqa} and answer sentence selection \cite{yu2014deep_trec}. Sometimes, it is beneficial to leverage feature engineering heuristics when combined with statistical approaches. The feature engineering heuristics or the external features are used to aid the learning model to successfully converge to a global solution \cite{QA1,yang2015wikiqa,talosintheswen}. The external features includes common observations such as number of n-grams, number of words match between headline and the body, cosine similarity between the headline and the body vector, etc. The FNC-1 baseline also includes a combination of feature engineering heuristics that alone achieves a competitive performance, even outperforming several widely used deep learning architectures. In this paper, we combine external features introduced in the baseline with some more heuristics that have been shown to be successful in other NLP tasks.

These days it is common to use pre-trained word embeddings such as Word2vec \cite{mikolov2013efficient} and GloVe \cite{pennington2014glove} along with deep models for NLP tasks. Similar to word embedding, the recurrent models have been used to encode an entire sentence to a vector. Some of the widely used sentence-to-vector models include doc2vec \cite{mikolov2013distributed}, paragraph2vec \cite{para} and skip-thought vectors \cite{kiros2015skip}. These deep recurrent models helps to capture the semantic and contextual information of the textual pairs, in our case, body and its claim. In our work, we use the skip-thought vector to encode the headline and the body, and combine it with external features and statistical approaches. 

Finally, the main contributions of the paper can be summarized as 
\begin{enumerate}
\item We propose an approach that is based on the combination of statistical, neural and feature engineering heuristics which achieves state-of-the-art performance on the task of fake news identification.
\item We evaluate the proposed approach on FNC-1 challenge, and compare our results with the top-4 submissions to the challenge. We also analyze the applicability of several state-of-the-art deep models on FNC-1 dataset.
\end{enumerate}

The rest of the paper is organized as follows. In section 2, we brief the previous idea over which our works builds, which is followed by applicability of state-of-the-art deep architectures on the problem of stance detection. In section 4 we describe the proposed approach in detail, followed by the experiment setup in section 5, that includes dataset description, training parameters, evaluation metrics used and results. Finally, our work is concluded in section 6.

\section{Related Work}
In this section, we discuss some previous work that is in relation to fake news identification such as rumor detection in news articles and hoax news identification. We also discuss the use of deep learning architecture used by some of the researchers with whom our work shares some similarity. 

\textbf{Fake news}. From an NLP perspective, researchers have studied numerous aspects of credibility of online information. For example, \cite{castillo2013predicting} applied the time-sensitive supervised approach by relying on the tweet content to address the credibility of a tweet in different situations.  \cite{chen2017call} used LSTM in a similar problem of early rumor detection. In an another work, \cite{chen2017ikm} aimed at detecting the stance of tweets and determining the veracity of the given rumor with convolution neural networks. A submission \cite{augenstein2016usfd} to the SemEval 2016 Twitter Stance Detection task focuses on creating a bag-of-words auto encoder, and training it over the tokenized tweets.

\textbf{FNC-1 submissions}. In their work, \cite{pfohlstance} achieved a preliminary score of 0.8080, slightly above the competition baseline of 0.7950. They experimented on four basic models on which the final result was evaluated: Bag Of Words (BOW), basic LSTM, LSTM with attention and conditional encoding LSTM with attention (CEA LSTM). In our work, instead of using the models separately, we combine the best of these models.

Another team, \cite{talosintheswen}, combined multiple models in an ensemble providing 50/50 weighted average between deep convolution neural network and a gradient-boosted decision trees. Though this work seems to be similar to our work, the difference lies in the construction of ensemble of classifiers. In a similar attempt, a team \cite{athenefnc}
concatenated various features vectors and passed it through an MLP model.

The work by \cite{riedel2017simple}, focuses on generating lexical and similarity features using (TF-IDF) representations of bag-of-words (BOW)  which are then fed through a multi-layer perceptron (MLP) with one hidden layer. In their work, \cite{chaudhrystance} divided the problem into two groups: \textit{unrelated} and \textit{related}. They were able to achieve ~90\% accuracy on the related/unrelated task by finding maximum and average Jaccard similarity score across all sentences in the article and choosing appropriate threshold values. A similar work of splitting the problem into two subproblems (\textit{related} and \textit{unrelated}) is also performed by \cite{chopra2017towards}. The work by \cite{millerfake} focuses on the use of recurrent models for fake news stance detection. 

\section{Technique Used}
\subsection{Deep Learning Architectures}
\label{sec_bidirectional}

To predict the stance for a given sample in FNC-1 dataset, a multi-channel deep neural network can be used to encode a given headline-body pair, which can be classified into one of the four stances. This is achieved by using a multi channel convolution neural network with $softmax$ layer at the output (shown in Figure 1). Similarly, instead of using the convolution and pooling layers, LSTM and GRU can be used to encode the headline-body pairs. The LSTMs and GRUs encode the given sequence of words into fixed length vector representation which can be used to score the relevance of headline-body pair. However, for long sequences, such as the body of a news article (which typically contain hundreds of words),
the RNN models fail to completely encode the entire information into a fixed length vector. A solution to this problem is given in the form of attentional mechanism \cite{learning_phrase_rep} which computes a weighted sum of all the encoder units that are passed on to the decoder. The decoder is learned in such a way that it gives importance to only some of the words.
The attention mechanism also alleviates the bottleneck of encoding input sequences to fixed length vector and have been shown to outperform other RNN based encoder-decoder models on longer sequences \cite{attn1}. To alleviate the problem of limited memory we use attention mechanism as described in \cite{attn1}.

We experiment with some of the deep architectures that have been shown to be successful in NLP tasks (shown in Figure 1). Most of these architectures have been proven to be effective for non-factoid based question answering \cite{all_arc1,all_arc2}.

\begin{figure*}[h!]
\centering\includegraphics[width=1\textwidth,height=0.25\textheight]{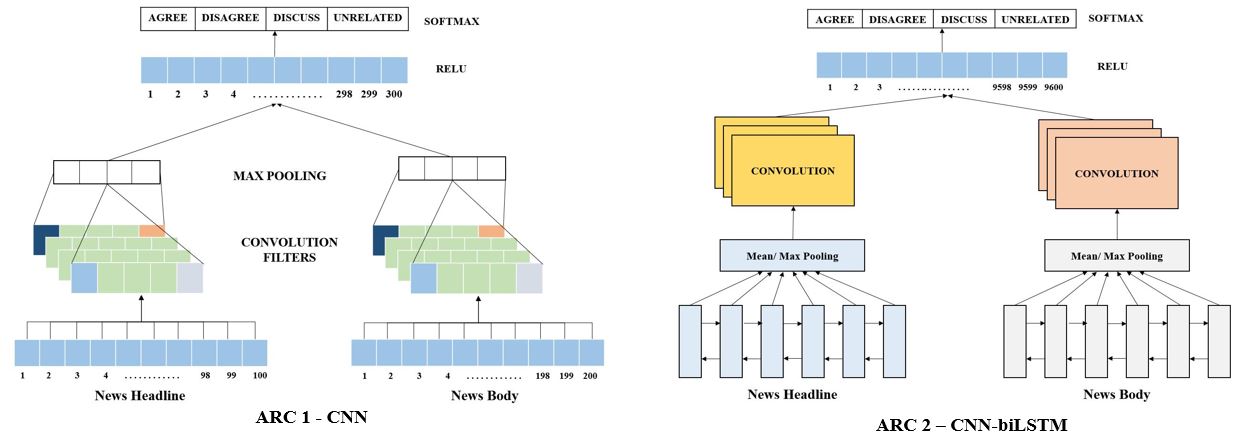}
\caption{Deep architectures used for stance detection on the FNC-1 dataset.}
\label{fig:one}
\end{figure*}

\section{Proposed Idea}

The $unrelated$ headline-body pairs in the FNC-1 dataset are created by randomly assigning a news body to the given headline. This type of data augmentation has been successfully used in NLP problems such as non-factoid question answering where it results in reasonable performance by the deep learning models \cite{QA1,QA2}. However, in the case of FNC-1 challenge, the $agree$, $disagree$, and $discuss$ headline-body pairs are relatively smaller in quantity than the $unrelated$ stance. This bias leads to a uneven distribution of dataset across the four classes, with the $unrelated$ category being the least interesting. Interestingness of a headline-body pair is evaluated in terms of information that it contains; It is easier to evaluate a $unrelated$ pair, while the other three are contingent on exploring contextual relationship between the headline and its body, and are considered more interesting.

The uneven distribution of FNC-1 dataset thwarts the performance of deep learning architectures introduced in Section 3. Moreover, news articles are heavily influenced by some words that are generally associated with news to describe its polarity. For example, words like $crime$, $accident$, and $scandal$ are often used with negative connotation. If such words are present in both the news headline, or are present in one while absent from the other, then, it is easier to identify such a pair as $agree$ or $disagree$. Deep learning models are dependent on a huge training corpus (few million headline-body pairs) in order to identify such nuances in patterns. The FNC-1 dataset, though the largest publicly available dataset on stance detection, does not satiate this criteria. For this reason, we introduce a much simpler strategy that consists of heavy use of feature engineering. We leveraged several widely used state-of-the-art features used in natural language processing, and use a feed-forward deep neural network which aggregates all the individual features and computes a score for each headline-body pair.

\subsection{Neural Embeddings}
We use skip-thought vectors which encodes sentences to vector embedding of length 4800 (shown in Figure 2).
\begin{figure*}[t!]
\centering\includegraphics[width=1\textwidth,height=0.25\textheight]{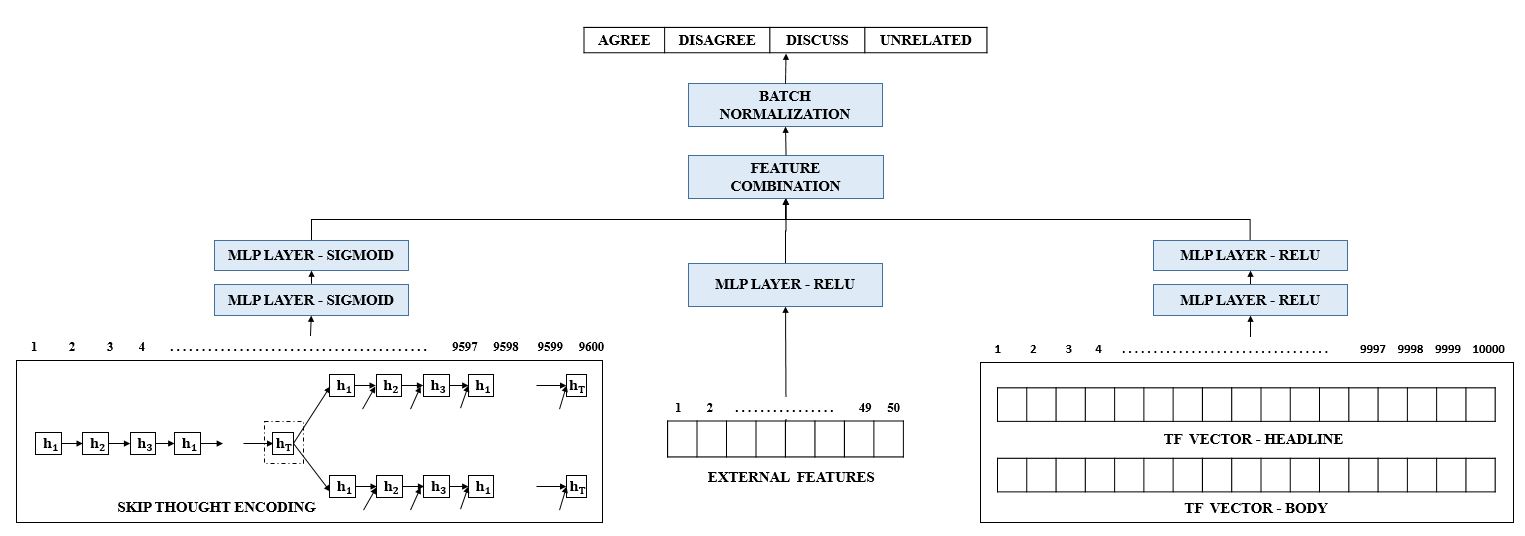}
\caption{Combining the neural, statistical and external features using deep MLP.}
\label{fig:two}
\end{figure*}
The skip-thought \cite{kiros2015skip} is a encoder-decoder based recurrent model that computes the relative occurrence of sentences.  In our work, we use the pre-trained skip-thought embedding which is trained on BookCorpus \cite{book}. We make the use of a pre-trained model since the FNC-1 dataset is relatively smaller than the dataset required to efficiently train a recurrent encoder-decoder model like skip-thought. 

We follow the work of \cite{kiros2015skip,sick} and compute two features from the skip-thought embeddings. These features have been shown to be effective in evaluating contextual similarity between sentences. The task of stance detection is analogous to the computation of contextual similarity between two sentences - headline and its body. We speculate that the features introduced by \cite{kiros2015skip,sick} should be effective for stance detection as well. Given the skip-thought encoding of news and headline as $u^{news}$ and $v^{head}$, we compute two features

\begin{align}
feat_1 &= u^{news} . v^{head} \\
feat_2 &= |u^{news} - v^{head}|
\end{align}

where $feat_1$ is the component-wise product and $feat_2$ is the absolute difference between the skip-thought encoding of news and headlines. Both of these features results in a 4800 dimensional vector each.

\subsection{Statistical Features}
We capture the statistical information from the text to vectors with the help of BOW, TF-IDF and n-grams models. We follow the work of \cite{riedel2017simple} and \cite{davisfake_stanford}, and produce the following vectors for each headline-body pair
\begin{enumerate}
\item 1-gram TF vector of the headline.
\item 1-gram TF vector of the body.
\end{enumerate}
This gives us a vector of 5000 dimension each. We concatenate both of the TF vectors and pass it to a MLP layer (as shown in Figure 2).

\subsection{External Features}
The external features include feature engineering heuristics such as number of similar words in the headline and body, cosine similarity between vector encodings of headline-body pairs, number of n-grams matched between the pairs, etc. We leveraged ideas for computing the external features from the baseline and add some extra features, which includes
\begin{enumerate}
\item Number of characters n-grams match between the headline-body pair, where $n =2,\cdots,16$.
\item Number of words n-grams match between the headline-body pair, where $n =2,\cdots,6$.
\item Weighted TF-IDF score between headline and its body using the approach mentioned in \cite{yu2014deep_trec}.
\item Sentiment difference between the headline-body pair, also termed as polarity and is computed using lexicon based approach.
\item N-gram refuting feature which is constructed using BOW on a lexicon of $n$ pre-defined words. It is similar to polarity based features with an addition of n-gram model.
\end{enumerate}
All the external features adds up to a 50-dimensional feature vector and is passed to a MLP layer similar to neural and statistical features. 

\section{Experimentations}
\subsection{Dataset Description}
We use the dataset provided in the FNC-1 challenge which is derived from the Emergent Dataset \cite{ferreira2016emergent}, provided by the fake news challenge administrators. The former consist of 49972 tuple with each tuple consisting of a headline-body pair followed by a corresponding class label \textit{stance} of either \textit{agree}, \textit{disagree}, \textit{unrelated} or \textit{discuss}. Word counts roughly ranges between 8 to 40 for headlines and 600 to 7000 for article body.
The distribution of FNC-1 dataset is shown in Table 2.

\begin{table}[h]
\centering
\begin{tabular}{|l|l|l|l|l|}
\hline
News articles & unrelated & discuss & agree     & disagree  \\ \hline
49972 & 73.13 \% & 17.83 \% & 7.36 \% & 1.68 \% \\ \hline
\end{tabular}
\caption{FNC-1 dataset description.}
\end{table}

\begin{table*}[t!]
\centering
\begin{tabular}{|c|c|c|c|}
\hline
\textbf{Hyperparameter} & \textbf{Skip-thought} & \textbf{External Features} & \textbf{TF-IDF Vectors}\\
\hline
\hline
MLP layers         	& 2 & 1 & 2                          \\
MLP neurons        	& 500 ; 100 & 50 & 500 ; 50             \\ 
Dropout   			& 0.2 ; - & - & 0.4 ; -           \\ 
Activation 			& sigmoid ; sigmoid & relu & relu ; relu \\
Regularization   & L2 - 0.00000001 ; - & - & L2 - 0.00005 ; -\\ 
\hline
\hline
MLP Layers & \multicolumn{3}{c|}{1}      \\
MLP neurons        	&  \multicolumn{3}{c|}{4}       \\
Activation			& \multicolumn{3}{c|}{Softmax}  	\\
Optimizer           & \multicolumn{3}{c|}{Adam}      \\
Learning rate           & \multicolumn{3}{c|}{0.001}      \\
Batch size           & \multicolumn{3}{c|}{100}      \\
Loss			& \multicolumn{3}{c|}{Cross-entropy}   \\
\hline
\end{tabular}
\caption{Values of hyper-parameters. The first half of the table shows the parameters used in architectures for extracting individual features. The second half shows the parameter setting of the feature combination layer that is shown in Figure 2.}
\label{tab_hyperparameter}
\end{table*}

The final results are evaluated over a test dataset provided by fake news organization consisting of 25413 samples.

\subsection{Training parameters}
As shown in Figure 2, the proposed model computes the feature vectors separately and then combine these with the help of a MLP layer. We use cross-entropy as the loss function to optimize our architecture with a softmax layer at the output which classify the given headline-body pair into \textit{agree}, \textit{disagree}, \textit{discuss}, and \textit{unrelated}. The hyper-parameter setting is shown in Table 3. 

\subsection{Baselines and compared methods}
Organizers of FNC-1 have provided a baseline model that consists of a gradient-boosting classifier over n-gram subsequences between the headline and the body along with several external features such as word overlap, occurrence of sentiment using a lexicon of highly-polarized words (like \textit{fraud} and \textit{hoax}). With this simple yet elegant baseline it is possible to outperform some of the highly used deep learning architectures that we have used in our work. Following the work of \cite{pfohlstance}, we also introduce three new baselines for the FNC-1 dataset: word2vec+external features baseline, skip-thought baseline, and TF-IDF baseline. All these baselines focuses on performance of neural, statistical, and external features, when used individually.

We compare our proposed approach with the submissions of top 4 teams at FNC-1 \footnote{http://www.fakenewschallenge.org/ \\ https://competitions.codalab.org/competitions/16843\#results}, which includes the work by \cite{talosintheswen}, \cite{pfohlstance}, \cite{athenefnc} and \cite{riedel2017simple}. Apart from the top submissions at FNC-1, we also compare the proposed architecture with four deep learning architectures introduced in Section 3, namely, CNN, biLSTM, BiLSTM+Attention and CNN+biLSTM.

\subsection{Evaluation metrics}

From Table 2 it is evident that the FNC-1 dataset shows a heavy bias towards unrelated headline-body pairs. Recognizing this data bias and the simpler nature of the $related/unrelated$ classification problems, the organizers of FNC-1 introduced the following weighted accuracy score as their final evaluation metric.

\begin{align}
Score_1 &= Accuracy_{Related,Unrelated} \\
Score_2 &= Accuracy_{Agree, Disagree, Discuss}\\
Score_{FNC} &= 0.25*Score_1 + 0.75*Score_2
\end{align}

We use the $Score_{FNC}$ as the main evaluation criteria while comparing the proposed model with other related techniques. We also use the class-wise accuracy  for further evaluation of the performance of all the techniques.

\subsection{Results}

The results on FNC-1 test dataset are shown in Table 4. The first part of the table shows the performance of the baselines used in our work. The FNC-1 baseline achieves a score of $75.2$ which is better than the performance of all deep architectures introduced in Section 3. The FNC-1 baseline is comprised of training gradient tree classifier on the hand crafted features (described in Section 4.3). Provided the simplicity of this baseline, it is indeed remarkable to achieve such a high score. The FNC-1 baselines achieves $approx\ 7\%$ higher class-wise accuracy on \textit{unrelated} stance as compared to skip-thought baseline, whereas the latter receiving a higher $Score_{FNC}$. Skip-thought baselines achieves a higher accuracy on $agree$ and $discuss$ than the $unrelated$ stance. 
\begin{table*}[h!]
\centering
\begin{tabular}{c|c|cccc|c}
    \toprule
    Method & $Score_{FNC}$ & Agree & Disagree & Discuss & Unrelated & Overall\\
    \midrule
    FNC-1 baseline & 75.20 & 9.09 & 1.00 & 79.65 & 97.97 & 85.44\\
    Word2vec + External Features & 75.78 & 50.70 & \textbf{9.61} & 53.38 & 96.05 & 82.79\\
    Skip-thought baseline & 76.18 & 31.8 & 0.00 & 81.20 & 91.18 & 82.48\\
    TF-IDF baseline & 81.72 & 44.04 & 6.60 & 81.38 & 97.90 & 88.46\\
    \midrule
    SOLAT in the SWEN \cite{talosintheswen} & 82.05 & 58.50 & 1.86 & 76.18 & 98.70 & 89.08\\
    Athene \cite{athenefnc} & 81.97 & 44.72 & 9.47 & 80.89 & \textbf{99.25} & \textbf{89.50}\\
    UCL Machine Reading \cite{riedel2017simple} & 81.72 & 44.04 & 6.60 & 81.38 & 97.90 & 88.46\\
   	Chips Ahoy! \cite{chips} & 80.12 & 55.96 & 0.28 & 70.29 & 98.98 & 88.01\\
    \midrule
    CNN & 60.91 & 35.89 & 2.10 & 46.77 & 88.47 & 74.84\\
    biLSTM & 63.11 & 38.04 & 4.59 & 58.13 & 78.27 & 69.88\\
    biLSTM + Attention & 63.17 & 58.74 & 0.03 & 63.48 & 77.49 & 73.27\\
    CNN + biLSTM  & 64.95 & \textbf{74.09} & 2.46 & 57.85 & 74.87 &  72.89\\
     Proposed & \textbf{83.08} & 43.82 & 6.31 & \textbf{85.68} & 98.04 & 89.29 \\
    \bottomrule
  \end{tabular}
\caption{Performance of different models on FNC-1 Test Dataset. The first half of the table shows the baselines, followed by the top-4 submissions, and different architectures used in our work. Column 2-5 shows the class-wise accuracy in \% while the last column shows the overall accuracy.}
\label{tab:res_semeval}
\end{table*}
Since the interestingness of $agree$ and $discuss$ is higher than the $unrealted$ stance, therefore, skip-thought achieves a higher $Score_{FNC}$. This also explains the reason for the introduction of new scoring criterion by the FNC organizers (see Section 5.4). Finally, the $Score_{FNC}$ by skip-thought, external features, and TF-IDF baselines are higher than the FNC-1 baseline. Therefore, our speculation to combine these three baselines models, is guaranteed to achieve a higher score on $Score_{FNC}$ evaluation metric. Moreover, all the baselines achieves very low or zero score on the \textit{disagree} stance. Therefore, apart from the $Score_{FNC}$, the class-wise performance is worth considering as a performance criterion.

The performance of top-4 teams that participated in FNC-1 are shown in the middle part of Table 4, with $SOLAT\ in\ the\ SWEN$ \cite{talosintheswen} winning the challenge achieving a score of 82.05. All the teams achieved higher score and class-wise accuracy on all stances except for the $disagree$ stance. This should be a concern, since the importance of $disagree$ is equivalent to the $agree$ and $discuss$ stance. We observed that the news pairs in the $disagree$ category are not only very few, but also consists of divergent news articles. This is one of the reason for poor performance of most of the deep models, including the top teams, on identifying $diagree$ stance.

The lowest section in Table 4 shows the performance of the proposed model along with other architectures used in our work. The proposed model achieves highest score and highest class-wise accuracy on \textit{discuss} stance whereas achieving high accuracy on other stances that is comparable to top submissions at FNC-1. From Table 5, it is evident that the overall accuracy achieved by the proposed model is slightly lower than \cite{athenefnc}, although the proposed model outperformed all the other techniques by a clear margin (in terms of $Score_{FNC}$). The possible reason for this deviation is that the \cite{athenefnc} gives more focus to the classification of $unrelated$ stances rather than the rest, which is the reason for highest overall accuracy. Since $unrelated$ stances are of least interest to us, this results in lower $Score_{FNC}$.
\begin{table*}[t!]
\centering
\begin{tabular}{c|cccc|c}
    \toprule
     & Agree & Disagree & Discuss & Unrelated & Overall\\
    \midrule
    Agree & 834 & 15 & 945 & 109 & 43.82\\
    Disagree & 208 & 44 & 328 & 117 &  6.31\\
    Discuss & 401 & 23 & 3825 & 215  & 85.68\\
    Unrelated & 22 & 12 & 325 & 17990 & 98.04\\
    \bottomrule
  \end{tabular}
\caption{Confusion matrix for the proposed model on FNC-1 testset.}
\label{tab:res_semeval}
\end{table*}
Finally, a confusion matrix is given in Table 5 that provides in-detail analysis of the performance of our approach.

\section{Conclusion}
In this paper, we explore the benefit of incorporating neural, statistical and external features to deep neural networks on the task of fake news stance detection. We also presented in-depth analysis of several state-of-the-art recurrent and convolution architectures (shown in Figure 1). The presented idea leverages features extracted using skip-thought embeddings, n-gram TF-vectors and several introduced hand crafted features.

We found that the uneven distribution of FNC-1 dataset undermines the performance of most deep learning architectures. The fewer training samples adds further to this aggravation. Creating a dataset for a complex NLP problems such as fake news identification is indeed a cumbersome task, and we appreciate the work by the FNC organizers, yet, a more detailed and elaborate dataset should make this challenge more suitable to evaluate.

\bibliography{refs}
\end{document}